\documentclass{article}
\usepackage{iclr2027_conference,times}
\iclrfinalcopy

\usepackage{hyperref}
\usepackage{url}
\usepackage{booktabs}
\usepackage{amssymb}
\usepackage{algorithm}
\usepackage{algpseudocode}
\usepackage{graphicx}
\usepackage{caption}
\usepackage{array}
\usepackage{tabularx}
\usepackage[table]{xcolor}
\usepackage{tcolorbox}
\tcbuselibrary{breakable,listings}

\newtcblisting{promptbox}[1]{
  title={#1},
  breakable,
  colback=gray!3,
  colframe=gray!55,
  colbacktitle=gray!12,
  coltitle=black,
  fonttitle=\bfseries\small,
  boxrule=0.5pt,
  arc=2pt,
  left=6pt,right=6pt,top=5pt,bottom=5pt,
  listing only,
  listing options={
    language={},
    basicstyle=\ttfamily\footnotesize,
    columns=fullflexible,
    keepspaces=true,
    breaklines=true,
    breakatwhitespace=false,
    showstringspaces=false,
    literate={“}{{\textquotedblleft}}1 {”}{{\textquotedblright}}1
             {ℝ}{{\ensuremath{\mathbb{R}}}}1
             {≥}{{\ensuremath{\geq}}}1
             {←}{{\ensuremath{\leftarrow}}}1,
    tabsize=2
  }
}

\newcommand{\aks}{OS-Wizard}
\title{\aks{}: Scaling Computer-Using Agents \\ to Specialized Software \\ via Knowledge-Guided Exploration}
\author{\textbf{Zhoumianze Liu}$^1\ ^2$
\textbf{Zichen Ding}$^2$
\textbf{Bowen Yang}$^5$
\textbf{Zehao Li}$^5$
\textbf{Zhenyu Wu}$^6$ \\
\textbf{Yibo Zhao}$^4$
\textbf{Kanzhi Cheng}$^7$
\textbf{Yian Wang}$^7$
\textbf{Yanan Sun}$^2$
\textbf{Qiushi Sun}$^3\ ^\dagger$
\textbf{Yu Qiao}$^2\ ^\dagger$\\
$^1$Fudan University
$^2$Shanghai AI Laboratory
$^3$The University of Hong Kong \\
$^4$East China Normal University
$^5$University of Science and Technology of China \\
$^6$Shanghai Jiaotong University
$^7$Nanjing University
$^8$National University of Singapore \\
\text{zmzliu23@m.fudan.edu.cn} \ 
\text{qiushisun@connect.hku.hk} \ 
\text{qiaoyu@pjlab.org.cn}
}

\usepackage{amsmath,amsfonts,bm}

\def\eqref#1{equation~\ref{#1}}

\def\1{\bm{1}}

\DeclareMathAlphabet{\mathsfit}{\encodingdefault}{\sfdefault}{m}{sl}
\SetMathAlphabet{\mathsfit}{bold}{\encodingdefault}{\sfdefault}{bx}{n}

\begin{document}
\maketitle
\pagestyle{plain}
\thispagestyle{plain}

\begin{abstract}
Computer-using agents are increasingly capable of performing complex tasks in the digital world. To advance such agents, supervised learning on annotated interaction trajectories has become a common approach to improving their task-solving capabilities. However, high-quality interaction trajectories remain scarce for specialized desktop applications, where limited application knowledge, complex GUI workflows, and incomplete exploration hinder automated collection. To address these challenges, we introduce \aks{}, a knowledge-guided automated trajectory collection framework. It combines tutorial-guided exploration with a hybrid GUI--Code action space, providing application-specific knowledge and simplifying GUI workflows through direct execution of application-native commands. The framework also checkpoints and revisits pivotal states during tree exploration to explore alternative workflows, complementing linear rollouts for broader coverage. We collect 10.9K automatically validated trajectories and use them to fine-tune open-source base models. The resulting model achieves 64.1\% accuracy on a benchmark of specialized desktop workflows, improving performance by 58.2 percentage points, establishing a new state of the art. The fine-tuned model also improves performance on general desktop workflows using GUI actions only, suggesting that the collected supervision benefits general desktop tasks as well. Comparative experiments and ablations demonstrate the efficiency of our collection framework and the value of code actions in specialized workflows.
\end{abstract}

\section{Introduction}

Computer-using agents (CUAs) extend foundation models from understanding multimodal information to directly interacting with computers, enabling them to complete tasks through graphical and programmatic interfaces across web and desktop environments~\citep{deng2023mind2web,zhou2024webarena,xie2024osworld,yuan2026osworld2}. Given a task instruction and observations such as screenshots, a CUA repeatedly reasons about the current state and executes actions such as mouse clicks or keyboard input~\citep{hong2024cogagent,zheng2024gpt,agashe2025agent}. As these agents are increasingly expected to handle longer and more complex workflows, their capabilities depend critically on diverse and high-quality interaction trajectories that provide supervision for both interface grounding and multi-step decision making.

However, computer-using trajectories remain substantially more difficult to obtain than conventional text or vision data, and existing datasets still rely heavily on costly human demonstrations~\citep{wang2026opencua,liu2026scalecua,qin2025ui}. Recent work has begun to automate trajectory synthesis through model-driven exploration and task generation~\citep{xu2025agenttrek,sun2025genesis,wei2026anchor, cheng2026openmobile}, but collecting useful data for specialized desktop applications faces three challenges. First, the scarcity of application-specific interaction data leaves general-purpose models with limited knowledge of how to use these applications correctly, hindering effective exploration. Second, even when applications expose domain-specific languages (DSLs), issuing commands through the GUI requires additional interface interactions, such as locating and clicking the appropriate input fields, introducing failure points beyond those of direct command execution. Third, current trajectory collection often explores applications through isolated interaction sequences, which may overlook important application capabilities and fail to fully exploit promising intermediate states.

\begin{figure}[t]
    \centering
    \includegraphics[width=\linewidth]{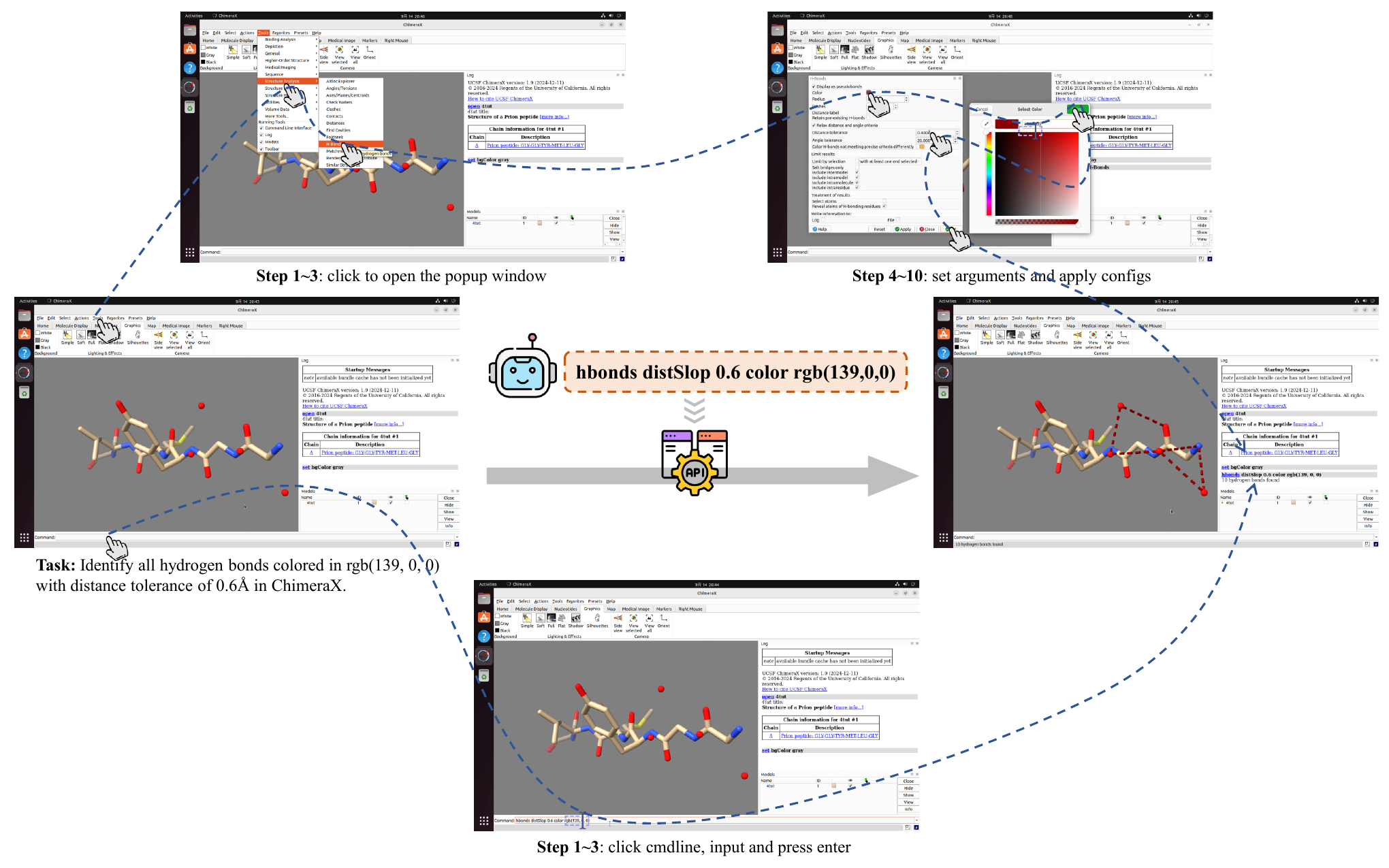}
    \caption{\textbf{Three interaction paths for the same visualization task in ChimeraX.} \textbf{Top:} A GUI-only workflow takes 10 actions to open the dialog, configure parameters, and apply the settings. \textbf{Middle:} Our domain-specific code interface executes the equivalent application-native command with a single action. \textbf{Bottom:} Entering the command through the GUI takes 3 actions to click the input, type the command, and press Enter.}
    \label{fig:intro}
\end{figure}

To address these challenges, we introduce \textbf{\aks{}}, an automated trajectory collection framework for specialized desktop applications. \aks{} augments GUI interactions with domain-specific code actions that directly submit application-native commands, including DSL expressions, reducing the interface interactions required to execute expert operations. In Figure~\ref{fig:intro}, this abstraction reduces a multi-step hydrogen-bond visualization workflow to a single code action. To equip the collector with the knowledge needed to use this action space, we retrieve application documentation and tutorials that explain application workflows and command usage, guiding exploration and task execution in unfamiliar software. Beyond independent linear rollouts, we introduce tree exploration that efficiently revisits \emph{pivotal states} during trajectory collection to explore diverse and feasible execution paths, thereby improving functional coverage.

We evaluate \aks{} on ScienceBoard~\citep{sun2026scienceboard}, a benchmark of specialized scientific applications, and OSWorld~\citep{xie2024osworld}, a general desktop benchmark. On ScienceBoard, fine-tuning Qwen3.5-27B on our collected trajectories improves overall task success from 5.9\% to 64.1\%, surpassing the reported overall human performance of 59.8\% and establishing a new state of the art. Under a matched 10K-sample training budget, \aks{} achieves 36.5\% success compared with 13.0\% and 18.9\% for the Task-driven and Branch-driven collection baselines, demonstrating that our collection framework produces more effective supervision per sample. On OSWorld, the fine-tuned model improves success from 39.1\% to 42.9\% using GUI actions only, which shows that the collected supervision supports substantial gains on specialized applications while also improving general desktop task performance.

Our main contributions are:
\begin{itemize}
\item We introduce \aks{}, an automated trajectory collection framework for specialized desktop applications, using tutorials to mitigate knowledge gaps and revisiting pivotal states to explore alternative workflows beyond independent linear rollouts.
\item We develop a validatable hybrid GUI--Code action space that reduces the interface interactions needed to execute application-native commands, with ablations supporting the value of code supervision.
\item We collect 10,919 trajectories comprising 54,938 state--action samples, improving performance on ScienceBoard and OSWorld and outperforming alternative collection strategies under a matched training-data budget.
\end{itemize}

\section{Related Works}
\paragraph{Computer-using Agents.}
Existing computer-using agents primarily emulate human computer use through GUI interactions. Cradle~\citep{tan2024cradle}, CogAgent~\citep{hong2024cogagent}, SeeClick~\citep{cheng2024seeclick}, Agent S~\citep{agashe2025agent}, UFO~\citep{zhang2025ufo}, and UI-TARS~\citep{qin2025ui} use mouse and keyboard actions to perform tasks from visual observations. Recent work extends this paradigm with programmatic interfaces: CoAct-1~\citep{song2025coact} and ActionEngine~\citep{zhong2026actionengine} introduces coding actions for executable Python or Bash programs; UltraCUA~\citep{yang2025ultracua}, UFO$^2$~\citep{zhang2025ufo2}, and Axis~\citep{lu2025axis} incorporate high-level tools or native application APIs; and ToolCUA~\citep{hu2026toolcua} and EE-MCP~\citep{he2026ee} study hybrid GUI--tool interaction and orchestration through MCP-based interfaces. These systems focus on general-purpose computer use, with limited dedicated support for the domain-specific knowledge and operations required by specialized desktop applications.

\vspace{-0.1cm}
\paragraph{Data Synthesis for GUI Agents.}
The scarcity of high-quality interaction trajectories has motivated scalable data generation for GUI agents. Tutorial-based methods such as AgentTrek~\citep{xu2025agenttrek}, TongUI~\citep{zhang2025tongui}, and Video2GUI~\citep{xiong2026video2gui} convert web tutorials into executable trajectories, while task-driven approaches, including AgentSynth~\citep{xie2026agentsynth} and Insta~\citep{trabucco2025insta}, synthesize task intents and execute them with capable agents. 
Exploration-driven methods such as OS-Genesis~\citep{sun2025genesis}, UI-ReWalk~\citep{lin2025gui}, OpenMobile~\citep{cheng2026openmobile, yan2026openmobile}, and AutoPlay~\citep{ramrakhya2026scaling} instead derive tasks or workflows from environment interaction. 
Recent systems further improve structure and scalability: ANCHOR~\citep{wei2026anchor} expands verified demonstrations from meaningful branch states, VeriCUA~\citep{sun2026veri} channels agentic exploration for scalable verification, and SEE~\citep{fan2026see} models GUI state transitions for long-horizon synthesis. However, dedicated mechanisms for exploring domain-specific operations and collecting verified trajectories in specialized desktop applications remain underexplored.

\vspace{-0.1cm}
\paragraph{Advancing GUI Agents with External Knowledge.}
Agentic retrieval augments conventional RAG with iterative planning, query refinement, and tool-based retrieval, allowing agents to adapt the retrieval process to the task~\citep{zhang2024agentic,jin2025search,singh2025agentic,du2026rag,suresh2026agenticrag}. Domain- and document-grounded agents apply these mechanisms to specialized knowledge, with PaperQA~\citep{lala2023paperqa}, RAGentA~\citep{besrour2025ragenta} and SQuAI~\citep{besrour2025squai} combining multi-agent retrieval and evidence verification for attributed and scientific QA, while LongAgent~\citep{zhao2024longagent}, M3DocAgent~\citep{cho2024m3docrag}, RAG-Anything~\citep{guo2025rag}, VisRAG~\citep{yu2025visrag}, and Patho-AgenticRAG~\citep{zhang2026patho} ground multimodal and long-document reasoning in external references. These methods primarily ground reasoning and answer generation in retrieved knowledge, leaving its role in application-specific actions and workflow exploration underexplored.

\begin{figure}[t]
    \centering
    \includegraphics[width=\linewidth]{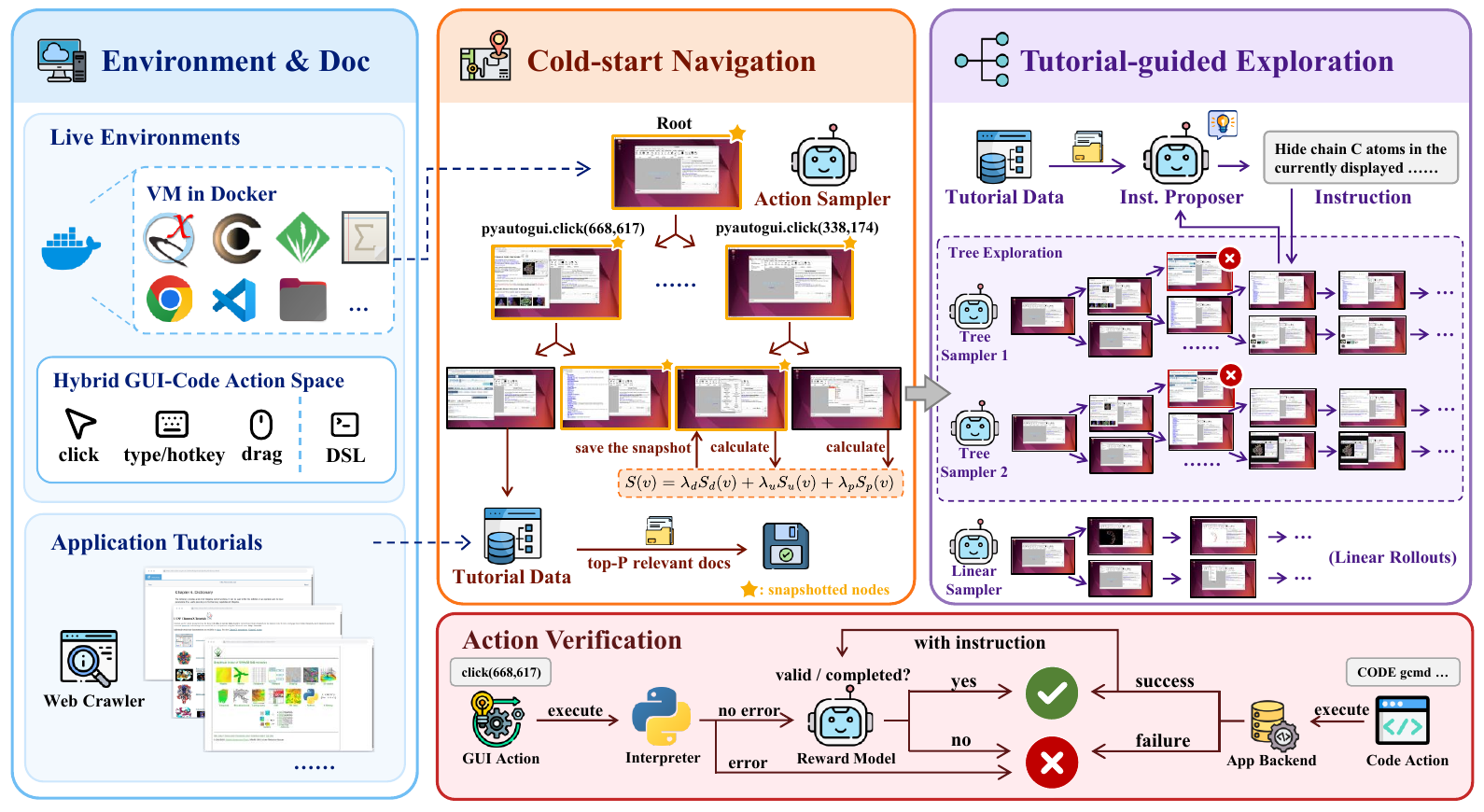}
    \caption{\textbf{Overview of trajectory collection in \aks{}.} \textbf{Left:} Tutorials guide exploration with GUI and code actions. \textbf{Center:} Cold-start navigation builds an initial tree and checkpoints pivotal states. \textbf{Right:} Parallel tree exploration and linear rollouts collect trajectories. \textbf{Bottom:} Execution checks and reward-model verification ensure trajectory quality.}
    \label{fig:method}
\end{figure}

\section{\aks{}}

\aks{} is an automated framework for collecting computer-using agent trajectories in specialized applications. Its key insight is to explore professional software through a hybrid GUI--Code action space, capturing both graphical interactions and domain-specific operations. To scale data collection over this action space, \aks{} combines structured tree exploration with complementary linear rollouts. Figure~\ref{fig:method} illustrates both collection strategies: tree exploration builds on cold-start navigation and pivotal-state checkpointing to support tutorial-guided parallel exploration, while linear rollouts collect independent task-oriented trajectories. Both strategies combine execution checks with stage-specific verification.

We model desktop GUI exploration as a rooted exploration tree $T=(V,E)$ with maximum depth $D$, where each node $v=(s, x, t) \in V$ represents a distinct visited state instance, consisting of a screenshot $s$, accessibility information $x$, and a textual state description $t$. Each edge $e=(v,a,v') \in E$ records an action $a \in A$ executed from node $v$ that non-deterministically produces a new child state instance $v'$. Linear rollouts arise as a special case of this formulation: when every non-root node is restricted to have at most one child, the exploration tree becomes a collection of independent linear trajectories sharing the same initial root state.

\subsection{Pivotal State Identification}
\label{sec:pivot}

Naively expanding GUI state trees incurs space complexity of
$O(N^D)$ for branching factor $N$, while repeatedly recovering successive states along a $d$-step trajectory via action replay under BFS incurs a cumulative cost of $O(d^2)$. Since GUI states differ substantially in their downstream exploration value, we define a sparse set of \emph{pivotal states} as states expected to serve as high-value branching points for subsequent exploration. \aks{} allocates more exploration budget to such states and stores VM snapshots for them, reducing both redundant branching and recovery overhead.

Given adjustable weights $\lambda_d$, $\lambda_u$ and $\lambda_p$ and candidate set of states $C$, we score each state $v \in C$ by
\begin{equation}
S(v) = \lambda_d S_d(v)+\lambda_u S_u(v)+\lambda_p S_p(v),
\end{equation}
where each term is min--max normalized before aggregation. $S_d(v)$ is a depth prior favoring shallow states, which expose broader functional regions and are shared by more trajectories. Specifically, the root is always treated as a pivot. $S_u(v)$ measures the magnitude of the preceding state transition based on changes in both the screenshot and the textual state description. Larger transitions such as page-level changes are more likely to introduce useful branching contexts. $S_p(v)$ optionally incorporates empirical action success rate from $v$ when such statistics are available, favoring states with reliable interactions. In practice, we select the top $\lceil 0.2|C| \rceil$ states in $C$ by $S(v)$ as pivots.

\subsection{Two-stage Tree Exploration}
\label{sec:two_stage}

\begin{figure}[t]
\centering
\begin{minipage}[c]{0.58\linewidth}
\begin{algorithm}[H]
\caption{Two-stage collection in tree exploration}
\label{alg:two_stage}
\scriptsize
\setlength{\baselineskip}{8pt}
\algrenewcommand\algorithmicindent{0.8em}
\algrenewcommand\alglinenumber[1]{\scriptsize\makebox[1.8em][r]{#1:}}
\algtext*{EndIf}
\algtext*{EndWhile}
\algtext*{EndFor}
\begin{algorithmic}[1]
\Statex \textbf{Stage I: Cold-start Navigation}
\State $V_1\gets\textsc{Expand}(v_0,N),\quad V_2\gets\bigcup_{v\in V_1}\textsc{Expand}(v,N)$
\State $S \gets \textsc{Pivot}(V_1,V_2),\quad \textsc{Snapshot}(S)$
\State $\textsc{Retrieve}(v,P)\ \forall v\in V_2$

\Statex \textbf{Stage II: Tutorial-guided Parallel Exploration}
\ForAll{workers \textbf{ in parallel}}
    \State $Q\gets$ randomly sparsified $V_2$ with $S$ preserved
    \While{$Q\neq\varnothing$}
        \State $(v,d)\gets\textsc{Pop}(Q)$
        \If{$f(d)>1$}
            \State $Q\gets Q\cup
            \textsc{Verify}\!\left(\textsc{Expand}(v,f(d))\right)$
        \Else
            \State $\textsc{Rollout}\bigl(v,\textsc{Synth}(\text{tutorials},\text{exemplars}),D\bigr)$
        \EndIf
    \EndWhile
\EndFor
\end{algorithmic}
\end{algorithm}
\end{minipage}\hfill
\begin{minipage}[c]{0.40\linewidth}
\centering
\includegraphics[width=\linewidth]{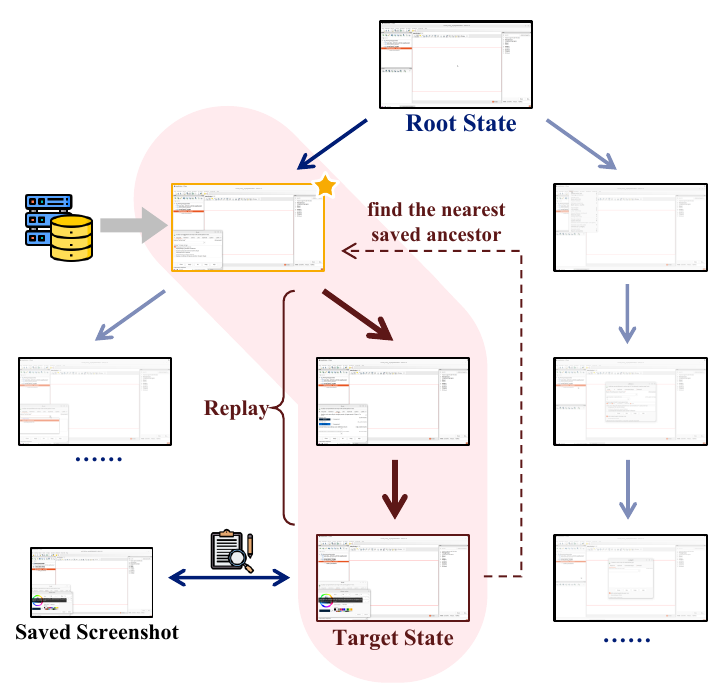}
\end{minipage}
\caption{\textbf{Left:} Two-stage collection in the tree exploration component of \aks{}. \textbf{Right:} Snapshot-assisted state recovery (\S\ref{sec:recover}): find and restore the nearest saved ancestor, then replay all actions along the path to the target while checking screenshot similarity.}
\label{fig:collection_recovery}
\end{figure}

The tree exploration component of \aks{} proceeds in two stages: \emph{Cold-start Navigation} and \emph{Tutorial-guided Parallel Exploration}, as is shown in Algorithm~\ref{alg:two_stage}. Exploration follows a depth-dependent branching function $f(d)=\max\{N-3d, 1\}$, where $N=10$ denotes the root branching factor. We truncate exploration beyond the maximum depth $D=16$. The motivation and design of this branching schedule are discussed in \S\ref{sec:budget}.

\paragraph{Cold-start Navigation.}
Starting from the root state, a single worker performs open-ended exploration, where an LLM sampler proposes diverse and non-redundant actions from the current screenshot and exploration history. The worker expands $f(0)=N$ first-level states and oversamples each of them to $N$ second-level states. Pivotal states are then identified using the scores in \S\ref{sec:pivot}: each first-level state forms a singleton candidate set and is thus retained as a pivot, while each group of ten second-level siblings forms a candidate set from which $2$ pivots are selected and checkpointed. For every second-level state, we additionally retrieve the top-$P$ relevant tutorials with $P=5$.

\paragraph{Tutorial-guided Parallel Exploration.}
Each worker receives the same cold-start tree but independently sparsifies and expands it. Within each group of second-level siblings, we randomly drop $M=N-f(1)$ nodes while always preserving pivots, reducing the effective branching factor to $f(1)$. The remaining nodes are inserted into a BFS queue and expanded according to $f(d)$. For a pivotal state, sampling continues until obtaining up to $f(d)$ valid children, whereas a non-pivotal state is attempted at most $f(d)$ times and may therefore yield fewer children. At branching states with $f(d)>1$, we use the same open-ended action sampling as in cold-start navigation to diversify action-level expansions. Once $f(d)$ decreases to $1$, the retrieved tutorials and 5--10 OOD exemplar instructions guide the synthesis of a task instruction encompassing both the exploration prefix and the subsequent rollout. With tutorials as context, the worker follows the instruction until completion or the maximum depth is reached. The single branch is executed continuously without BFS recovery between intermediate states. Independent dropout and action sampling lead workers toward different branches, improving trajectory diversity and depth. Newly generated transitions are validated according to their action types before being added to the tree as described in \S\ref{sec:action}.

\subsection{Validatable Hybrid Action Space}
\label{sec:action}

The key design of \aks{} is its hybrid action space $A=A_{\mathrm{GUI}}\cup A_{\mathrm{Code}}$. Professional applications often expose command-line interfaces or domain-specific languages alongside graphical controls, making GUI-only interaction unnecessarily long or brittle for many expert operations. $A_{\mathrm{GUI}}$ contains PyAutoGUI-based interactions, while $A_{\mathrm{Code}}$ contains domain-specific command primitives. Domain-specific code actions do not expose privileged functionality, but abstract commands that could otherwise be entered and executed through the GUI. They may be provided directly by the application or benchmark, or realized by automatically entering the corresponding commands through the GUI based on accessibility information. See \S\ref{sec:code} for code actions supported by each application and additional implementation details.

Quality control is integrated into action generation and post-execution verification. Each sampled action must be grounded in the current screenshot and accessibility information and, when an instruction is present, remain consistent with the task objective. At branching states, previously sampled sibling actions are provided as negative references, and duplicate or semantically redundant actions are rejected to promote intent-level diversity. Recent execution history is also included to suppress local action loops.

All actions must execute without errors. Without a task instruction, GUI actions additionally require reward-model validation of transition validity, whereas code actions are accepted if the application reports no execution error. With an instruction, both action types undergo execution checks, and the reward model judges only task completion: success terminates the rollout; otherwise, execution continues until completion or the depth limit $D$. Trajectories terminated solely by this limit are excluded. Lean reference-proof replay uses pre-collection compilation checks and a separate completion procedure (\S\ref{sec:lean}).

\subsection{Snapshot-assisted State Recovery}
\label{sec:recover}

To reduce BFS recovery cost and improve robustness to non-deterministic GUI behavior, \aks{} checkpoints all pivotal states in $S \subseteq V$. As illustrated in the right part of Figure~\ref{fig:collection_recovery}, given a target state $v$, we recover from its nearest checkpointed ancestor
\begin{equation}
p(v)=\operatorname*{arg\,max}_{u\in \operatorname{Anc}(v)\cap S} d(u),
\end{equation}
by loading the snapshot of $p(v)$ and then replaying only the actions along the path from $p(v)$ to $v$. Since QXL graphics state cannot be reliably restored from QEMU snapshots, we use an Xvfb-based virtual desktop instead; see Appendix~\S\ref{sec:vm} for details.

During the replay, each recovered state is verified against its recorded screenshot via visual similarity. A failed step is retried up to $R = 10$ times before the recovery is aborted. For applications with time-varying or network-dependent content, we adopt a relaxed visual-consistency threshold to tolerate benign rendering differences.

\section{Experiments}
\subsection{Experimental Settings}

\paragraph{Benchmarks.}
We conduct experiments on ScienceBoard~\citep{sun2026scienceboard} and OSWorld~\citep{xie2024osworld}. Both benchmarks are adapted to the Qwen3 tool-call format~\citep{bai2025qwen3}, where ScienceBoard uses the hybrid GUI--Code action space described in \S\ref{sec:action} for all applications except Celestia, which lacks a programmatic control interface comparable to the command-line or DSL available in other applications, while OSWorld uses GUI actions only.

\paragraph{Data Collection.}
To balance collection quality and cost, we use GPT-5.6-Sol~\citep{openai2026gpt56}, GPT-5.5~\citep{openai2026gpt55}, and Kimi-K2.5~\citep{team2026kimi} as instruction proposers and action samplers, GPT-5.6-Sol as the reward model for transition and completion verification, and Qwen3-Embedding-8B~\citep{zhang2025qwen3} for tutorial retrieval. We collect data using both \aks{} tree exploration and linear rollouts. The resulting corpus contains 10,919 trajectories and 54,938 state--action training samples, averaging 5.03 steps per trajectory, including terminal steps. Tree exploration contributes 32,250 samples, while linear rollouts contribute 22,688 samples. Of the full corpus, 17,442 samples (31.75\%) come from general-purpose applications, while the remaining 37,496 samples come from specialized applications. Lean uses only linear rollouts with an application-specific collection procedure. Detailed collection statistics and the Lean-specific procedure are provided in \S\ref{sec:data} and \S\ref{sec:lean}.

\paragraph{Baselines.}
For ScienceBoard, we compare against (1) base models, (2) proprietary models, (3) some related results from other open-source models, and (4) SFT models trained on human-collected, \emph{Task-driven}, and \emph{Branch-driven} trajectories. We implement the latter two following OS-Genesis~\citep{sun2025genesis} and ANCHOR~\citep{wei2026anchor}. For OSWorld, where our goal is to assess whether \aks{} training preserves and improves general desktop capability, we primarily compare against the corresponding Qwen3.5-27B base model.

\paragraph{Training Setup.}
We perform full-parameter supervised fine-tuning of Qwen3-VL-8B~\citep{bai2025qwen3} and Qwen3.5-27B~\citep{qwen3.5} on the collected trajectories using Qwen3-style tool-call conversations. Training configurations, conversation construction, and evaluation details are provided in \S\ref{sec:train}.

\begin{table}[t]
\centering
\footnotesize
\renewcommand{\arraystretch}{1.05}
\resizebox{\linewidth}{!}{
\begin{tabular}{l c c c c c c c}
\toprule
\textbf{Model} & \textbf{Astron} & \textbf{Biochem} & \textbf{GIS} & \textbf{Alg} & \textbf{ATP} & \textbf{Doc} & \textbf{Overall} \\
\midrule

Human$^\dagger$
& 51.5
& 69.0
& 55.9
& \underline{74.2}
& \textbf{42.3}
& 68.8
& \underline{59.8} \\
\midrule

\multicolumn{8}{l}{\emph{Base models}} \\
Qwen3-VL-8B~\citep{bai2025qwen3}
& 12.1
& 10.3
& 2.9
& 9.7
& 0.0
& 0.0
& 6.5 \\

Qwen3.5-27B~\citep{qwen3.5}
& 12.1
& 13.8
& 2.9
& 0.0
& 0.0
& 6.3
& 5.9 \\

\midrule
\multicolumn{8}{l}{\emph{Proprietary models}} \\
GPT-5.5~\citep{openai2026gpt55}
& 18.2
& 41.4
& 11.8
& 9.7
& 3.8
& 31.3
& 18.3 \\

GPT-5.6-Sol~\citep{openai2026gpt56}
& 36.4
& 69.0
& \underline{58.8}
& 41.9
& 3.8
& 37.5
& 42.6 \\

Claude-Opus-4.8~\citep{anthropic2026opus48}
& 24.2
& 65.5
& 8.8
& 35.5
& 0.0
& 37.5
& 27.8 \\

Gemini-3.1-Pro~\citep{googledeepmind2026gemini31pro}
& 18.2
& 6.9
& 2.9
& 22.6
& 0.0
& 25.0
& 11.8 \\

\midrule
\multicolumn{8}{l}{\emph{Open-source models}} \\
Qwen3-VL-235B-A22B~\citep{bai2025qwen3}
& 9.1
& 31.0
& 0.0
& 9.7
& 0.0
& 6.25
& 9.5 \\

Qwen3.5-397B-A17B~\citep{qwen3.5}
& 9.1
& 17.2
& 5.9
& 6.5
& 0.0
& 6.3
& 7.7 \\

Qwen3.8-27B~\citep{qwen3.8}
& 36.4
& 65.5
& 41.2
& 61.3
& 0.0
& 43.8
& 42.0 \\

Kimi-K2.5~\citep{team2026kimi}
& 3.0
& 27.6
& 0.0
& 9.7
& 0.0
& 12.5
& 8.3 \\

ScaleCUA-32B~\citep{liu2026scalecua}$^\dagger$
& 12.1
& 10.3
& 0.0
& 9.7
& 0.0
& 0.0
& 5.9 \\

SCALECUA-Qwen3.5-9B~\citep{lv2026scalecua}$^\dagger$
& \textbf{60.6}
& 62.1
& 35.3
& 67.7
& 19.0
& \underline{86.7}
& 54.0 \\

\midrule
\multicolumn{8}{l}{\emph{Sampling methods}} \\
Human Data
& 12.1
& 24.1
& 0.0
& 9.7
& 0.0
& 12.5
& 9.5 \\

Task-driven~\citep{sun2025genesis}
& 15.2
& 13.8
& 14.7
& 19.4
& 0.0
& 12.5
& 13.0 \\

Branch-driven~\citep{wei2026anchor}
& 21.2
& 20.7
& 20.6
& 29.0
& 0.0
& 18.8
& 18.9 \\

\midrule
\multicolumn{8}{l}{\emph{Ours}} \\

\aks{}-Qwen3-VL-8B
& 36.3$_{\pm 2.5}$
& \underline{71.2}$_{\pm 1.6}$
& 50.0$_{\pm 4.2}$
& 54.8$_{\pm 2.6}$
& 6.4$_{\pm 1.8}$
& 64.5$_{\pm 7.8}$
& 46.6$_{\pm 0.6}$ \\

\aks{}-Qwen3.5-27B-10K
& 20.2$_{\pm 1.4}$
& 51.7$_{\pm 0.0}$
& 43.1$_{\pm 2.8}$
& 65.6$_{\pm 4.0}$
& 6.4$_{\pm 1.8}$
& 20.8$_{\pm 3.0}$
& 36.5$_{\pm 0.7}$ \\

\aks{}-Qwen3.5-27B
& \underline{52.5}$_{\pm 1.4}$
& \textbf{83.9}$_{\pm 1.6}$
& \textbf{68.6}$_{\pm 2.8}$
& \textbf{77.4}$_{\pm 2.6}$
& \underline{21.8}$_{\pm 1.8}$
& \textbf{87.5}$_{\pm 5.1}$
& \textbf{64.1}$_{\pm 0.7}$ \\

\bottomrule
\end{tabular}}
\caption{
ScienceBoard success rates (\%).
$^\dagger$ denotes results reported in the original paper; all other results are evaluated by us with a maximum step limit of 50.
Results in the \emph{Ours} block are reported as mean $\pm$ standard deviation over three evaluation runs, while results of SCALECUA-Qwen3.5-9B are reported as means.
Best results are \textbf{bolded} and the second best results are \underline{underlined}.}
\label{tab:main_sb}
\end{table}

\subsection{Main Results}

\paragraph{ScienceBoard.}
Table~\ref{tab:main_sb} reports success rates on ScienceBoard. \aks{}-Qwen3.5-27B achieves state-of-the-art overall performance among the compared methods at 64.1\%, surpassing the reported SCALECUA-Qwen3.5-9B result of 54.0\% by 10.1 percentage points and the reported human performance of 59.8\%. Across all individual domains, \aks{}-Qwen3.5-27B ranks first or second among the compared models and reported human results. Across both backbones, fine-tuning on \aks{} trajectories yields substantial improvements: Qwen3-VL-8B improves from 6.5\% to 46.6\%, while Qwen3.5-27B increases from 5.9\% to 64.1\%. Compared with the newer Qwen3.8-27B base model at 42.0\%, Qwen3.5-27B starts from a much lower success rate, indicating limited task-execution capability in these specialized domains before fine-tuning. Its subsequent improvement highlights the effectiveness of the collected supervision in developing this capability. Notably, even \aks{}-Qwen3-VL-8B surpasses the strongest evaluated proprietary model, GPT-5.6-Sol (46.6\% vs.\ 42.6\%), showing that our supervision can train a smaller model to outperform the strongest model used for data collection. Finally, under the same 10K-sample training budget, \aks{}-Qwen3.5-27B-10K substantially outperforms both our Task-driven and Branch-driven baselines (36.5\% vs.\ 13.0\% and 18.9\%), demonstrating the effectiveness of the overall collection pipeline under a matched data budget.

\paragraph{OSWorld.}
Figure~\ref{fig:main_anly}(a) reports success rates on OSWorld. Fine-tuning Qwen3.5-27B on \aks{} trajectories improves the success rate from 39.1\% to 42.9\%, a gain of 3.8 percentage points. The training corpus contains 17,442 samples from general-purpose applications, accounting for 31.75\% of all 54,938 samples, while the remaining samples come from specialized applications. These results show that training on this corpus improves general desktop performance alongside the gains on ScienceBoard.

\section{Analysis}

\begin{figure*}[t]
\centering
\begin{minipage}[c]{0.38\textwidth}
    \vspace{0pt}
    \centering
    \small
    \setlength{\tabcolsep}{4pt}
    \begin{tabularx}{\linewidth}{
        @{}>{\raggedright\arraybackslash}X r@{}
    }
        \toprule
        \textbf{Model} & \textbf{Overall} \\
        \midrule
        \multicolumn{2}{@{}l}{\textit{Base models}} \\
        Qwen3-VL-8B $^\dagger$ & 33.9 \\
        Qwen3.5-27B & 39.1 \\
        \midrule
        \multicolumn{2}{@{}l}{\textit{Ours}} \\
        \aks{}-Qwen3.5-27B & \textbf{42.9} \\
        \bottomrule
    \end{tabularx}
\end{minipage}

\hfill
\begin{minipage}[c]{0.58\textwidth}
    \vspace{0pt}
    \centering
    \small
    \setlength{\tabcolsep}{4pt}
    \begin{tabularx}{\linewidth}{
        @{}>{\raggedright\arraybackslash}X r r@{}
    }
        \toprule
        \textbf{Configuration} & \textbf{Samples} & \shortstack{\textbf{ScienceBoard}\textbf{(\%)}} \\
        \midrule
        \textbf{\aks{}} & Full & \textbf{64.1} \\
        \midrule
        \aks{} (mixed) & 20K & 55.0 \\
        \quad w/o code supervision & 20K & 26.6 \\
        \quad w/o linear rollouts & 20K & 41.4 \\
        \quad w/o tree exploration & 20K & 30.7 \\
        \bottomrule
    \end{tabularx}
\end{minipage}

\vspace{-0.5em}
\caption{\textbf{(a)} OSWorld success rates (\%).
\aks{}-Qwen3.5-27B uses a maximum step limit of 50, and $^\dagger$ denotes results reported in the original paper.
\textbf{(b)} ScienceBoard ablations on domain-specific code supervision and trajectory collection strategies using Qwen3.5-27B. The code ablation randomly samples trajectories containing no code calls throughout, yielding 20K training samples.}
\label{fig:main_anly}
\end{figure*}

\subsection{Ablations on Components}
\label{sec:ablation}

As shown in Figure~\ref{fig:main_anly}(b), we ablate domain-specific code supervision and trajectory collection strategies to understand where the gains of \aks{} come from. As a key design of the framework, the hybrid GUI--Code action space enables concise abstraction of expert operations in specialized applications. Under a matched budget of 20K state--action training samples, training on mixed data achieves 55.0\% ScienceBoard success, whereas training on data randomly sampled from trajectories containing no code calls achieves 26.6\%. This 28.4 percentage-point gap supports the value of including code supervision in the training corpus.

Tutorial guidance is particularly important for specialized applications where general-purpose models exhibit low task success. For example, GPT-5.5 achieves only 9.7\% and 11.8\% success rates on KAlgebra and GRASS GIS. In our no-document collection runs on these applications, 75.71\% of branches reach the maximum depth $D=16$ without completion and are discarded, substantially increasing the collection cost per usable trajectory and making data yield less reliable. This indicates that application-specific guidance is important for efficient long-horizon trajectory collection.

We further train models using 20K samples drawn exclusively from either the tree-based or the linear collection strategy. These two variants achieve 41.4\% and 30.7\% respectively. Both substantially outperform the corresponding base model, demonstrating that each collection strategy provides useful training supervision even with the same amount of training data. These results motivate combining the two collection strategies to construct the final training corpus.

\subsection{Data and Model Size Scaling}
\label{sec:scale}

\begin{figure*}[t]
\centering
\begin{minipage}[t]{0.50\textwidth}
    \vspace{0pt}
    \centering
    \includegraphics[width=\linewidth]{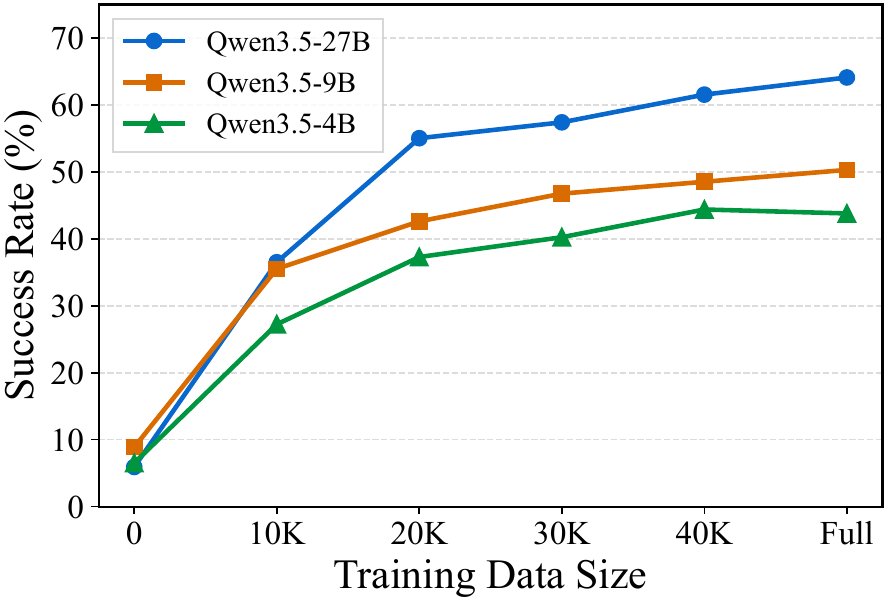}
\end{minipage}
\hfill
\begin{minipage}[t]{0.46\textwidth}
    \vspace{0pt}
    \centering
    \includegraphics[width=\linewidth]{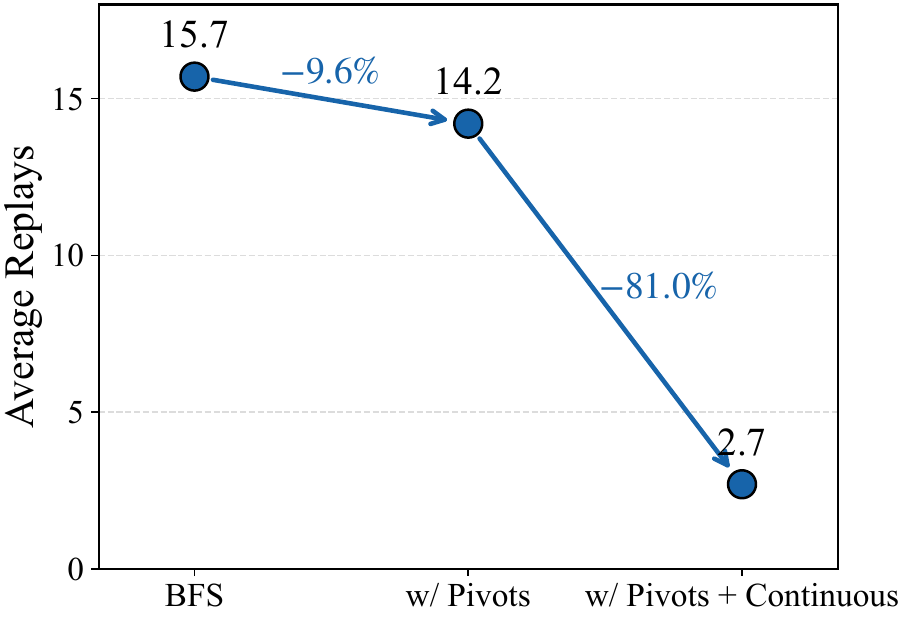}
\end{minipage}
\vspace{-0.5em}
\caption{Scaling and recovery-efficiency analysis of \aks{}.
\textbf{(a)} ScienceBoard success rate as a function of training data size for Qwen3.5-4B, 9B, and 27B.
\textbf{(b)} Average replay overhead of standard BFS, standard BFS with pivotal-checkpoint recovery, and standard BFS with pivotal checkpoints and continuous single-branch execution.}
\label{fig:anly}
\end{figure*}

\paragraph{Data Scaling.}
We first vary the number of state--action training samples while holding the backbone fixed. Figure~\ref{fig:anly}(a) shows that additional data generally improves ScienceBoard success rates, with substantial gains at small data scales and diminishing returns toward the full corpus. All three backbones improve markedly over their base models with the initial subset of training data, indicating that even limited \aks{} supervision is useful for adapting to specialized applications. The benefits also extend beyond this initial adaptation: for Qwen3.5-27B, expanding the training set from 10K samples to the full corpus yields a 27.6 percentage-point improvement. The returns to additional data nevertheless vary across backbones: performance continues to improve for the 9B and 27B models, while the 4B model largely plateaus at the upper end of the explored range. Thus, the results support expanding the collected corpus, while suggesting that the marginal benefit should be assessed for the target backbone.

\paragraph{Model Scaling.}
We next compare backbone sizes under the same training-data budget. Although the base models do not exhibit a monotonic ordering by size, larger backbones consistently achieve higher success rates at every non-zero data budget in Figure~\ref{fig:anly}(a). This change in ordering suggests that base-model performance alone does not fully reflect the benefits of model scale after application-specific fine-tuning. With the full corpus, scaling from 4B to 9B and from 9B to 27B improves the success rate by 6.5 and 13.8 percentage points respectively. Since these comparisons hold the amount of training data fixed, they identify backbone size as another effective axis for improving the trained agent. Within the evaluated Qwen3.5 family, a larger backbone can therefore deliver higher task success from an existing corpus, making model scale a useful consideration when the collection budget is fixed.

\subsection{Complexity Control}
\label{sec:budget}

The depth-dependent branching function $f(d)=\max\{N-\gamma d,1\}$ balances exploration breadth and trajectory depth while controlling tree growth. Since shallow states are shared by more downstream trajectories and often expose broader functional regions, we allocate more branching budget near the root and gradually reduce it with depth. Under $N=10$, $\gamma=3$, and $D=16$, the branching factors decrease as $10\!\rightarrow\!7\!\rightarrow\!4\!\rightarrow\!1$. With an action validation rate of 53.8\% estimated from a pilot run, approximately 187.6 branches per worker are expected to reach the single-branch instruction-following stage, preserving broad early coverage without continued combinatorial growth.

\aks{} reduces state-recovery overhead through pivotal checkpoints and continuous single-branch execution. In our data-collection runs, standard BFS requires replaying 15.7 actions on average, while recovering from pivotal checkpoints reduces this to 14.2. Once $f(d)=1$, branches are executed continuously without intermediate BFS recovery, further reducing the observed average to 2.7 replays. Together, these mechanisms control tree growth and the overhead of extending deep branches.

\section{Conclusion}

We introduced \aks{}, an automated trajectory collection framework for specialized desktop applications. \aks{} combines a validatable hybrid GUI--Code action space with tutorial-guided exploration and complementary tree-based and linear collection to generate diverse and useful interaction trajectories. Experiments on ScienceBoard and OSWorld show that training on the resulting data consistently improves computer-using agents, with particularly strong gains on specialized professional applications. These results demonstrate the potential of automated trajectory collection and richer action abstractions for scaling supervision beyond common GUI workflows. Future work may further strengthen trajectory quality control and explore more efficient collection architectures that better balance long-horizon exploration with state-storage and recovery costs. Another promising direction is to scale \aks{} to substantially larger data and model regimes.

\bibliography{iclr2027_conference}
\bibliographystyle{iclr2027_conference}

\newpage
\appendix
\section{Action Space and Code Definitions}
\label{sec:code}

Collection and evaluation share an OSWorld-style execution interface: GUI operations are expressed as PyAutoGUI commands, supplemented by application-specific \texttt{CODE} commands and task-control tokens. When constructing training trajectories, we converts collected actions to Qwen-style JSON tool calls; the trained model emits this same format at inference. The evaluation adapter then translates these calls back into executable environment actions.

In the model representation, the agent interacts through a single function, \texttt{computer\_use}, whose JSON arguments contain an \texttt{action} field and the corresponding parameters in Table~\ref{tab:action_space}. The interface exposes 14 GUI and task-control actions, plus an application-specific \texttt{code} action when enabled. ScienceBoard uses this hybrid interface except for Celestia\footnote{\url{https://celestiaproject.space/}}, which uses GUI actions only; OSWorld also uses GUI actions only.

\begin{table}[!hbp]
\centering
\caption{Model-facing action space in Qwen tool-call format and its environment execution semantics. Parameters are fields in the arguments of \texttt{computer\_use}; \texttt{coordinate} is an $(x,y)$ pair.}
\label{tab:action_space}
\footnotesize
\setlength{\tabcolsep}{4pt}
\renewcommand{\arraystretch}{1.15}
\begin{tabularx}{\linewidth}{@{}p{0.22\linewidth}p{0.21\linewidth}>{\raggedright\arraybackslash}X@{}}
\toprule
Action & Parameters & Execution semantics \\
\midrule
\texttt{key} & \texttt{keys} (list) & Press a single key or a key combination, releasing the keys afterwards. \\
\texttt{key\_down} & \texttt{key} (string) & Hold a key without releasing it. \\
\texttt{key\_up} & \texttt{key} (string) & Release a held key. \\
\texttt{type} & \texttt{text} (string) & Enter text using \texttt{pyautogui.typewrite}. \\
\texttt{mouse\_move} & \texttt{coordinate} & Move the cursor to the specified location. \\
\texttt{left\_click} & \texttt{coordinate} & Click the left mouse button. \\
\texttt{right\_click} & \texttt{coordinate} & Click the right mouse button. \\
\texttt{middle\_click} & \texttt{coordinate} & Click the middle mouse button. \\
\texttt{double\_click} & \texttt{coordinate} & Double-click the left mouse button. \\
\texttt{left\_click\_drag} & \texttt{coordinate} & Drag from the current cursor location to the target; default duration is 0.5 seconds. \\
\texttt{scroll} & \texttt{pixels} (number) & Pass the value to \texttt{pyautogui.scroll}; positive values scroll up and negative values scroll down. \\
\texttt{hscroll} & \texttt{pixels} (number) & Execute the same ordinary scroll operation as \texttt{scroll}. \\
\texttt{wait} & \texttt{time} (number) & Emit the environment's \texttt{WAIT} control token. \\
\texttt{terminate} & \texttt{status} & Emit \texttt{DONE} for \texttt{success} or \texttt{FAIL} for \texttt{failure}. \\
\texttt{answer} & \texttt{text} (string) & Submit an answer as \texttt{ANS <text>}. \\
\texttt{code} & \texttt{cmd} (string) & Dispatch \texttt{CODE <cmd>} to the application adapter. \\
\bottomrule
\end{tabularx}
\end{table}

The default coordinate convention is a normalized $1000\times1000$ grid. For a native screenshot of width $W$ and height $H$, the adapter maps $(x,y)$ to $(\lfloor xW/1000\rfloor,\lfloor yH/1000\rfloor)$. Click actions without coordinates operate at the current cursor position. Although the schema names the scroll argument \texttt{pixels}, the adapter forwards it directly to PyAutoGUI as a scroll amount without pixel conversion. Likewise, \texttt{wait} exposes a \texttt{time} argument but emits only \texttt{WAIT}, leaving the waiting duration to the environment. The \texttt{terminate} status is the agent's completion declaration, not an independently computed task-success score.

CODE actions provide application-specific command and file-editing primitives within the hybrid action space defined in \S\ref{sec:action}. Table~\ref{tab:code_definitions} lists their syntax and semantics, following the application interfaces. In a tool call, \texttt{cmd} contains the command without the \texttt{CODE} prefix; the evaluation adapter adds this prefix when dispatching the action. For example, arguments \texttt{\{"action":"code","cmd":"open 251d"\}} produce \texttt{CODE open 251d} for ChimeraX.

\begin{table}[t]
\centering
\caption{CODE action definitions. Angle brackets denote arguments supplied by the agent. File operations are shared by Lean 4 and TeXstudio; TeXstudio additionally supports forwarding commands to a running ChimeraX or KAlgebra instance.}
\label{tab:code_definitions}
\footnotesize
\setlength{\tabcolsep}{4pt}
\renewcommand{\arraystretch}{1.15}
\begin{tabularx}{\linewidth}{@{}p{0.14\linewidth}p{0.40\linewidth}>{\raggedright\arraybackslash}X@{}}
\toprule
Application & Syntax & Operation \\
\midrule
ChimeraX & \texttt{CODE <command>} & Execute a native ChimeraX command. \\
\midrule
GRASS GIS & \texttt{CODE gcmd <cmdline>} & Execute a console command and display its output in the Tool window. \\
& \texttt{CODE switch <location> <mapset>} & Switch to the specified location and mapset under \texttt{/home/user/grassdata}. \\
& \texttt{CODE load <layer>} & Load the named layer from the current mapset. \\
\midrule
KAlgebra & \texttt{CODE calc <expr>} & Submit an expression to the Calculator. \\
& \texttt{CODE 2d <expr>} & Add a two-dimensional curve. \\
& \texttt{CODE 3d <expr>} & Set the displayed three-dimensional surface. \\
\midrule
Editors & \texttt{CODE read <path>} & Read the complete file and return its text in the next message. \\
& \texttt{CODE overwrite <path> <text>} & Replace the entire file contents with the supplied text. \\
& \texttt{CODE insert <path> <index> <line>} & Insert text at the specified 1-based line position. \\
& \texttt{CODE replace <path> <src> <dst>} & Replace the first occurrence of the source string with the destination string. \\
& \texttt{CODE delete <path> <index>} & Delete the line at the specified 1-based position. \\
\midrule
TeXstudio & \texttt{CODE chimerax <command>} & Forward a native command to ChimeraX. \\
& \texttt{CODE kalgebra <operation> <arg>} & Forward a \texttt{tab}, \texttt{calc}, \texttt{2d}, or \texttt{3d} operation to KAlgebra with the semantics above. \\
\bottomrule
\end{tabularx}
\end{table}

For file-editing commands, paths cannot contain spaces, and only the final argument may contain spaces or newlines, except for \texttt{replace}, whose source and destination strings can each be enclosed in double quotes. Line indices start at 1. The \texttt{insert} operation strips leading and trailing whitespace from its text argument unless that argument is enclosed in double quotes, which preserves indentation. These quotes are argument delimiters and do not support escape sequences. The \texttt{overwrite} operation accepts multiline text, while \texttt{replace} changes only the first matching occurrence.

\paragraph{ChimeraX.}
We use the built-in RESTful remote-control interface of ChimeraX~\citep{pettersen2021ucsf}. The adapter submits the native command following the \texttt{CODE} prefix to this interface, allowing the agent to issue commands such as \texttt{CODE open 251d} directly to the running application.

\paragraph{GRASS GIS.}
We extend the wxGUI logic of GRASS GIS~\citep{neteler2012grass} with an interface for executing console commands. ScienceBoard~\citep{sun2026scienceboard} directly invokes \texttt{RunCommand} to execute instructions silently. Our implementation instead uses Python to simulate entering the command into the console and pressing the execution button, so that the command and its output appear on screen in the Tool window. This makes the execution output available in subsequent screenshots. The adapter also exposes operations for switching the active location and mapset and loading a named layer.

\paragraph{KAlgebra.}
KAlgebra provides calculator and 2D/3D plotting functionality~\citep{kde_kalgebra_handbook}. We extend the internal server introduced by ScienceBoard for KAlgebra. In addition to the existing 2D and 3D graph operations, we add \texttt{calc} to submit expressions to the Calculator. The adapter routes calculation and graph commands to the corresponding server endpoints, updating the running application.

\paragraph{Lean 4 and TeXstudio.}
We add file read/write interfaces for Lean 4~\citep{moura2021lean} and TeXstudio\footnote{\url{https://www.texstudio.org/}} without modifying either application. Both adapters support reading, overwriting, inserting, replacing, and deleting file contents, enabling direct edits to Lean source files and LaTeX documents. TeXstudio reuses the same file-operation implementation as Lean and additionally forwards prefixed ChimeraX and KAlgebra commands to their respective adapters when those applications are running. Both applications monitor file changes in real time and automatically reflect them in their respective editors.

\section{Implementation of VM}
\label{sec:vm}

Each environment runs QEMU~\citep{bellard2005qemu} (\texttt{qemu-system-x86\_64}) inside an Ubuntu 22.04 Docker container with access to the host's \texttt{/dev/kvm} for hardware acceleration. The guest also uses Ubuntu 22.04, installed with legacy BIOS boot. The default configuration allocates four virtual CPUs and 8\,GiB of RAM, with a 64\,GiB virtual disk in qcow2 format. QEMU uses the \texttt{pc-q35-4.2} machine type, the host CPU model, and VirtIO disk and network interfaces. The container entrypoint launches QEMU under \texttt{tini}; mounted directories \texttt{/storage} and \texttt{/qmp} hold disk images and the QEMU Machine Protocol (QMP) socket, respectively. The controller uses QMP to manage VM execution and checkpointing. The guest-side interaction server is accessible through forwarded port 5000, while port 5900 provides remote desktop access.

QXL with SPICE is available for initial configuration and debugging. For trajectory collection and state recovery, we launch QEMU with \texttt{-display none -vga none}, since QXL graphics state could not be reliably restored in our snapshot workflow. Inside the guest, Xvfb provides an X11 display at \texttt{:1} with a resolution of $1920\times1080$ and 24-bit color depth. GNOME Flashback with the Metacity window manager runs on this display, and \texttt{x11vnc} exposes it on port 5900. The desktop session is enabled through the \texttt{xtop} user service. This configuration places the virtual framebuffer and desktop processes in guest memory, allowing them to be captured together with application state without relying on QXL restoration. We enable toolkit accessibility and configure the guest for unattended interaction by disabling screen blanking, screen locking, and automatic sleep.

A checkpoint is a pair $C=(Q,M)$: $Q$ is a preserved qcow2 disk image, including its backing chain, and $M$ is a QEMU migration stream containing guest memory and execution state, including CPU and device state. Thus, the runtime snapshot is stored in a separate migration file. By default, disk writes use a copy-on-write overlay over a base image. Once an image becomes part of a checkpoint, subsequent execution writes to a new child overlay, preserving the checkpoint's disk state. A writable overlay belongs to a single running VM, while preserved backing images can be reused across restorations. The launcher also supports direct writes and temporary QEMU snapshot mode, but our checkpoint workflow uses explicit qcow2 overlays paired with migration files.

The VM lifecycle implements the state recovery in \S\ref{sec:recover} as follows:
\begin{enumerate}
    \item \textbf{Initial launch.} Create a writable overlay backed by the base qcow2 image, boot the guest using this overlay, and establish the QMP connection.
    \item \textbf{Save.} Pause the guest with QMP \texttt{stop} and export its runtime state to migration file $M$. After the export completes, terminate the current QEMU process and close the QMP connection. Preserve the current disk state as $Q$ by retaining its overlay and backing chain, producing checkpoint $(Q,M)$. Create a new writable overlay $O$ backed by $Q$, restart QEMU with $O$, and import $M$ to continue from the saved state.
    \item \textbf{Load.} Terminate the current VM and close its QMP connection, then discard its working overlay. For the requested checkpoint $(Q,M)$, create a fresh overlay $O'$ backed by $Q$ and start QEMU with the matching migration file $M$. The launcher imports this file through QEMU's \texttt{-incoming} option. Subsequent execution modifies only $O'$, allowing repeated exploration from the same checkpoint.
    \item \textbf{Final exit.} Terminate the VM, close the QMP connection, and discard its remaining working overlay. Retain saved checkpoint images, their backing chains, and migration files for later restoration.
\end{enumerate}
Pairing the preserved disk image with its corresponding migration file ensures that restored memory and disk state refer to the same checkpoint. After loading the nearest pivotal checkpoint, the collector replays the remaining actions and verifies recovered screenshots as described in \S\ref{sec:recover}.

Whenever a new VM is loaded, the controller registers cleanup handlers for normal program exit and termination signals such as \texttt{SIGINT}. A background polling loop additionally detects abandoned containers. Every \texttt{ACTIVE\_TIME} interval, it updates the current timestamp in \texttt{/var/log/heartbeat} inside the containers it manages and checks this file in containers matching the \texttt{os-aks} naming convention. Containers whose heartbeat age exceeds \texttt{INACTIVE\_TIME} are forcibly stopped. This complements exit handlers when a collector terminates without completing normal cleanup.

\section{Dataset statistics}
\label{sec:data}

The collected corpus contains 54,938 state--action training samples from 10,919 trajectories across 13 applications. Table~\ref{tab:dataset_statistics} reports the application-level breakdown. We count each serialized training record as one sample and each terminal record as the end of one trajectory. Trajectory lengths include the terminal step. Counts refer to the corpus before deduplication, including 173 exactly duplicated records. There are 9,862 distinct instruction strings, deduplicated within each application, and 36,226 distinct referenced screenshot paths; different trajectories may share instructions or screenshots.

\begin{table}[!hbp]
\centering
\caption{Application-level dataset statistics. Samples and mean trajectory lengths include terminal steps. The overall mean is weighted by trajectory count.}
\label{tab:dataset_statistics}
\small
\begin{tabular}{@{}lrrr@{}}
\toprule
Application & Trajectories & Training samples & Mean steps \\
\midrule
Celestia & 1,120 & 7,401 & 6.61 \\
ChimeraX & 1,411 & 6,458 & 4.58 \\
GRASS GIS & 1,561 & 5,487 & 3.52 \\
KAlgebra & 1,636 & 7,841 & 4.79 \\
Lean 4 & 1,486 & 7,546 & 5.08 \\
TeXstudio & 591 & 2,763 & 4.68 \\
\midrule
LibreOffice Calc & 428 & 2,676 & 6.25 \\
Chrome & 420 & 2,286 & 5.44 \\
GIMP & 409 & 2,279 & 5.57 \\
LibreOffice Impress & 455 & 2,541 & 5.58 \\
Nautilus & 514 & 2,522 & 4.91 \\
VS Code & 412 & 2,459 & 5.97 \\
LibreOffice Writer & 476 & 2,679 & 5.63 \\
\midrule
Total & 10,919 & 54,938 & 5.03 \\
\bottomrule
\end{tabular}
\end{table}

Tree exploration contributes 5,739 trajectories and 32,250 training samples, while linear rollouts contribute 5,180 trajectories and 22,688 samples, including the Lean proof trajectories described in \S\ref{sec:lean}. The six specialized applications contribute 37,496 samples. The remaining 17,442 samples (31.75\%) come from these general-purpose applications: LibreOffice, Chrome, GIMP, Nautilus, and VS Code.

Ordinary tree exploration and linear rollouts use a depth limit of $D=16$; Lean reference-proof replay is exempt. During conversion, paths with leaf depth greater than 40 are discarded before terminal training records are appended. In the resulting corpus, trajectory lengths, including terminal steps, have a mean of 5.03 and a median of 4; the 90th and 95th percentiles are 9 and 12, respectively, and the observed maximum is 40. Each training sample contains the current screenshot and up to four historical screenshots, with a mean of 3.15 images per sample. These repeated context images are counted separately from the distinct screenshot paths reported above.

Action statistics count only the target response of each training sample, excluding tool calls repeated in its conversation history. The corpus contains 55,471 target tool calls: 30,444 GUI operations, 13,037 code operations, and 11,990 task-control operations (\texttt{wait}, \texttt{answer}, or \texttt{terminate}). The processing pipeline also accepts responses with multiple tool calls, so call counts can exceed sample counts. Domain-specific code appears in 23.73\% of training samples and accounts for 29.98\% of GUI and code operations combined, excluding task-control operations from this denominator.

\section{Trajectories Collection for Lean 4}
\label{sec:lean}

Lean 4 poses additional challenges for automatic trajectory collection. Limited familiarity with Lean 4 syntax, tactics, and library conventions makes correct proof generation difficult for the collection models, and entering and editing proofs through a GUI adds further interaction overhead. Generating theorem statements from scratch also provides no assurance that the resulting tasks admit proofs within the available interaction budget. To efficiently build supervision for automated theorem proving (ATP), we therefore construct desktop interaction trajectories from existing Lean proofs. We obtain 1,608 problems with proof code from the Formal Math Autoformalization Dataset~\citep{agenticcommons_formalmath_2026} and use their proofs to guide deterministic linear rollouts.

Before collection, we compile each theorem together with its reference proof and retain only those that pass compilation. For each retained problem, we preserve the imports, declarations, and target theorem statement, and replace the proof body with an indented \texttt{sorry} placeholder. The theorem statement with this unfinished proof serves as the task instruction. The collector launches a VM with VS Code open on the Lean project and stages the problem in \texttt{/home/user/sci/Sci/Basic.lean}, recording the initial screenshot. This setup fixes both the task and its reference solution before collection, while the recorded trajectory shows how to construct that solution through the same file-editing code interface used during evaluation.

As a concrete example, the following shows a reference proof retained for replay after preprocessing:

\begin{promptbox}{Lean 4 Reference Proof}
import Mathlib

theorem div_add_div_pos_ge_two (a b : ℝ) (ha : 0 < a) (hb : 0 < b) : a / b + b / a ≥ 2 :=
by
  rw [ge_iff_le, ← sub_nonneg]
  have hab : (0 : ℝ) < a * b := mul_pos ha hb
  have key : a / b + b / a - 2 = (a - b) ^ 2 / (a * b) := by
    field_simp
    ring
  rw [key]
  exact div_nonneg (sq_nonneg _) hab.le
\end{promptbox}

The collector replays the nonempty proof lines in order. For the example above, its first action is \texttt{CODE replace <path> sorry <first-line>}; subsequent actions use \texttt{CODE insert <path> <index> <line>} at the corresponding 1-based line positions. Subsequent lines retain their indentation, including nested tactic blocks. In the example above, the first action replaces \texttt{sorry} with \texttt{rw [ge\_iff\_le, $\leftarrow$ sub\_nonneg]}. Later actions insert the positivity fact, the auxiliary algebraic identity and its nested proof, and the concluding tactics. Each action adds one source line, which need not constitute a complete tactic block or a complete proof on its own. After each edit, the collector waits four seconds for the editor to update, saves a screenshot, and records the command and waiting time as a transition. This captures intermediate desktop states throughout proof construction rather than only the completed source file.

After replay, we compile the resulting Lean file; only trajectories that pass compilation are marked complete and retained. Intermediate edits may leave proofs incomplete, so compilation is not required at every step. The retained paths are converted into Qwen-format training conversations using the procedure in \S\ref{sec:train}. For Lean, the converter also reconstructs the source file after each edit and includes its current contents in the corresponding observation, allowing the model to condition on both the screenshot and the partially constructed proof. The proof recorder leaves action descriptions and reasoning fields empty, so these demonstrations primarily supervise the file-editing tool calls. After compilation checks, replay, and trajectory filtering, the final dataset contains 1,486 proof trajectories and 7,546 step-level training samples, including terminal steps, from the original 1,608 problems.

\section{Training and evaluation details}
\label{sec:train}

We fully fine-tune Qwen3-VL-8B and Qwen3.5-27B on $8\times$ and $16\times$ H200 GPUs, respectively, using Qwen3-style tool-call conversations. For ScienceBoard applications with code support, domain-specific code actions are represented as \texttt{\{"action":"code", "cmd":"..."\}} arguments to \texttt{computer\_use}.

Training conversations are constructed from the collected exploration paths using \texttt{traj.py}, following the representation conversion in \S\ref{sec:code}. The converter parses supported PyAutoGUI calls through Python's abstract syntax tree and maps their function names and arguments to \texttt{computer\_use} actions. Native screenshot coordinates $(u,v)$ are normalized to $(\lfloor1000u/W\rfloor,\lfloor1000v/H\rfloor)$ for screenshot dimensions $(W,H)$; the optional relative-coordinate input mode instead scales unit coordinates by 1000. For example, \texttt{pyautogui.click(960,540)} on a $1920\times1080$ screenshot becomes \texttt{\{"action":"left\_click","coordinate":[500,500]\}}. The \texttt{CODE} and \texttt{ANS} prefixes become the \texttt{code} and \texttt{answer} actions with their payloads preserved. Each action is serialized as a JSON call to \texttt{computer\_use} inside \texttt{<tool\_call>} tags.

The converter selects paths whose task is marked complete and whose leaf depth does not exceed 40, omits actions that cannot be translated, and appends a successful \texttt{terminate} action paired with the final screenshot. Each step yields a conversation containing the task instruction, the current screenshot, and up to four recent screenshot--response pairs; older steps are summarized by their action descriptions. Application-specific responses include the recorded \texttt{Thought}, \texttt{Action}, and tool call, whereas the generic desktop mode uses only \texttt{Action} and the tool call. The resulting conversations are written to JSONL with their image references for supervised fine-tuning. At evaluation time, \texttt{qwen.py} translates model-generated calls back to the execution interface using coordinate quantization and the action-specific conventions in Table~\ref{tab:action_space}.

We report the training configuration for the main Qwen3.5-27B experiment using the full corpus. We perform full-parameter supervised fine-tuning with LLaMA-Factory~\citep{zheng2024llamafactory} using the state--action conversations described above. The run concatenates the selected training datasets and uses the \texttt{qwen3\_vl} conversation template with the \texttt{qwen} tool format. The template option \texttt{enable\_thinking} is set to \texttt{false}; the supervised response content follows the conversation construction described above. Training uses DeepSpeed~\citep{rasley2020deepspeed} and BF16 precision, with a maximum sequence length of 32,768 tokens. Table~\ref{tab:training_config} summarizes the settings for this main experiment.

\begin{table}[t]
\centering
\caption{Training settings for the main Qwen3.5-27B experiment on the full corpus. Other experiments use similar training settings with minor adjustments. The effective global batch size is the per-device batch size multiplied by the gradient accumulation steps and the number of data-parallel workers.}
\label{tab:training_config}
\small
\begin{tabular}{@{}lc@{}}
\toprule
Setting & Qwen3.5-27B \\
\midrule
Fine-tuning method & Full-parameter SFT \\
DeepSpeed ZeRO stage & 2 \\
Hardware & $16\times$ H200 \\
Precision & BF16 \\
Maximum sequence length & 32,768 \\
Per-device batch size & 1 \\
Gradient accumulation steps & 8 \\
Effective global batch size & 128 \\
Learning rate & $1\times10^{-5}$ \\
Learning-rate schedule & Cosine \\
Warmup steps & 100 \\
Training epochs & 1 \\
Validation split & None \\
Checkpoint saving & End of epoch \\
\bottomrule
\end{tabular}
\end{table}

This run uses DeepSpeed ZeRO stage 2, with partitioned all-gather, reduce-scatter, communication overlap, contiguous gradients, and round-robin gradient handling enabled. The all-gather and reduction bucket sizes are each set to $2\times10^9$ elements. Batch sizes, gradient accumulation, gradient clipping, and BF16 enablement are configured as \texttt{auto} for resolution by the training framework; the configuration does not specify optimizer or parameter offloading. The activation-checkpointing configuration enables activation partitioning and contiguous-memory optimization, disables CPU checkpointing and checkpoint-boundary synchronization, and leaves the checkpoint count unspecified. Activation-checkpoint profiling and wall-clock breakdown are disabled.

The main Qwen3.5-27B run starts from its pretrained backbone without resuming a training checkpoint. We train for one epoch, save model weights at the end of the epoch, and use the resulting checkpoint for downstream evaluation. No validation split is held out, and evaluation during training is disabled. With one sample per device and eight gradient accumulation steps, the 16-GPU run has an effective global batch size of 128.

We evaluate on ScienceBoard and OSWorld using the Qwen3-style tool-call interfaces described in the main text. ScienceBoard exposes the hybrid GUI--Code action space, with domain-specific code operations defined in \S\ref{sec:code}; Celestia uses GUI actions only. OSWorld uses GUI actions only. We report task success rates on both benchmarks.

At each interaction step, the evaluation adapter receives the task instruction and current desktop screenshot. Its default history window contains the four most recent observation--response pairs, including their screenshots and full model responses. Earlier steps are represented by a numbered list of their action descriptions, without their screenshots or full responses. The current screenshot is appended to this context, together with the latest execution output when supplied by the environment. For Lean, the current contents of \texttt{/home/user/sci/Sci/Basic.lean} are additionally included as authoritative source text when available in the observation. Historical execution outputs and historical Lean source text are not separately replayed in the conversation. The adapter's reset routine clears its stored history between tasks.

Screenshots are resized before being encoded as PNG images. The resizing routine approximately preserves the aspect ratio while rounding each dimension to a multiple of 32. It uses a maximum image area of $16\times16\times4\times12{,}800=13{,}107{,}200$ pixels and first limits the longest side to 8,192 pixels. Mouse coordinates are mapped back to the original desktop resolution using the normalized convention in \S\ref{sec:code}.

The system prompt identifies the application, states its operating constraints, and supplies the \texttt{computer\_use} schema. These constraints include preserving the active ChimeraX session, using a single GRASS GIS map display, avoiding parametric 2D plots in KAlgebra, preserving the Lean proof prefix and prohibiting \texttt{sorry}/\texttt{admit}, and avoiding unrelated edits in TeXstudio. The model is prompted to return a brief \texttt{Thought}, an \texttt{Action} description, and one JSON function call enclosed in \texttt{<tool\_call>} tags. The adapter parses the call and converts it into PyAutoGUI operations or environment control commands, including application code commands. Relative to the original Qwen adapter, this implementation adds application-specific prompting and code dispatch, held-key operations, explicit answer submission and failure termination, and execution-output and Lean-source feedback.

\section{Prompts for Sampling}
\label{sec:prompt}

The following templates are used for instruction proposal, action sampling, and transition/completion verification. Braced names denote runtime substitutions; doubled braces in the Python source are rendered as literal JSON braces here. Screenshots are attached as images alongside the user messages: one current screenshot for proposal and action sampling, and the before/after pair for verification. The optional \texttt{reference} field contains reference instructions, and \texttt{examples} supplies a JSON array with \texttt{count} entries. The \texttt{primitive} field supplies domain-specific code definitions. The \texttt{output} field appends the latest command output when available. Missing optional inputs follow the empty-string or \texttt{<ABSENT>} conventions in the implementation.

All actions must execute without errors. Without an instruction, the reward model checks GUI transition validity, while code actions require only error-free application execution. The reward model's \texttt{succ} field indicates GUI transition validity in this mode. With an instruction, it indicates task completion for both action types; \texttt{succ=false} means the rollout continues until completion or the depth limit.

\subsection{Instruction Proposal}

\begin{promptbox}{Instruction Proposal: System Prompt}
You are an autonomous goal-proposal agent operating on a Ubuntu desktop using a screenshot and optional documents.
Given a screenshot, history instruction and some documents (if provided), propose {count} plausible user goals (inst) that a human might want to achieve next.
All proposed goals should be based on what is visible in the screenshot or explicitly stated in the documents or instruction. Do not assume hidden state, background tasks, or unstated intent.

Goal rules
Each goal must be actionable, realistic, and meaningfully different from the others.
Do not invent UI elements, files, or workflows not supported by the inputs.
Newly generated goals must not be identical to, or trivial rephrasings of, any previously used history instruction.
Use clear references instead of vague ones, e.g. we prefer 'draw a curve of y=x^2+1' rather than 'draw any of quadratic curves'.

{reference}
Output format
You must output exactly the following JSON wrapped in backticks. Do not include any extra text outside the array.
```json
{examples}
```
\end{promptbox}

\begin{promptbox}{Instruction Proposal: User Prompt}
Current observation:

[SCREENSHOT]
The provided image is a screenshot of the current Ubuntu desktop state

[DOCUMENTATION]
{document}

[HISTORY INSTRUCTION]
{instruction}
\end{promptbox}

\subsection{Action Sampling}

\begin{promptbox}{Action Sampling: System Prompt}
You are an autonomous GUI interaction agent operating on a Ubuntu desktop via visual observation and accessibility information.

Your task is to propose ONE plausible next user action that a real human user might reasonably perform next, based on the information provided:
- a screenshot of the current desktop,
- an accessibility (a11y) tree describing visible UI elements,
- (optional) a software-related document,
- (optional) an instruction,
- (optional) a list of reference commands, representing commands that have already been checked.
- (optional) a list of history commands with description, representing commands that have already been executed.

You are not executing the action, only proposing it. Your proposal must be grounded strictly in what is visible or explicitly provided. Do not assume hidden UI state, background tasks, or an implicit goal unless it is clearly implied by the document or instruction.

Document & Instruction Handling:
If instruction is provided, the proposed action MUST follow instruction.
If instruction is not provided but doc is provided, the proposed action SHOULD be guided by and consistent with doc.
If neither doc nor instruction is provided, choose a conservative, broadly applicable exploratory action based only on the screenshot and a11y tree, favoring actions that expand future interaction possibilities (e.g., opening a menu, revealing additional panels or controls, focusing the main window, scrolling a primary panel).
Never invent steps, commands, UI elements, or workflows that are not supported by the screenshot, a11y tree, doc, or instruction.

Action Form & Primitives:
The action must be either a single pyautogui sentence or a single predefined code primitive. Do not chain actions, do not combine primitives, and do not include conditionals or explanations.
CODE primitives are first-class actions and should be strongly preferred when the document or instruction references command-line usage, system-level operations, or scripted workflows.
CODE primitives should also be preferred when a UI interaction would be fragile, slow, or ambiguous, or when a primitive better matches the intent expressed in doc or instruction.
The following describe the available code primitives specified to current desktop environments:
{primitive}

Environment & Timing Constraints:
The computer is slow and UI responses may take a long time, so acting too quickly is risky while waiting longer than necessary is acceptable.
Wait times must always be multiples of 15 seconds such as 15, 30, 45, or 60. Average safe waiting time is approximately 30 seconds.

Reference Command Rule:
A reference command entry includes the full command, its associated wait time, and whether it was accepted or rejected.
You MUST NOT output a command that is identical to any reference command, nor one that is only a trivial variation of it (e.g., same action with slightly different coordinates, timing, wait values, or other inconsequential parameter changes). Changing only coordinates, timing, wait, or selecting the same UI element in a slightly different way is not considered a different intent.
Your proposed command must represent a meaningfully different intent or interaction compared to all reference command entries; if no such meaningful action exists, choose a different UI element or interaction direction instead of repeating or slightly modifying a previous one.
Attention: some rejected reference commands may have failed only because the wait time was too short. If the action involves complex UI operations, loading, or data downloads, repeating the same command with a longer wait time is allowed in this specific case, but it should be used sparingly and only when there is strong reason to believe insufficient waiting was the cause of failure.

History Command Rules:
A history command entry includes all commands executed from the moment the software was launched up to the current state, and is provided strictly for reference.
Attention, history command represents previously executed commands, while reference command represents potential commands that can be executed from the current state and have not yet been executed.
You MUST carefully review recently executed commands and avoid reissuing actions that have already been performed.
If the same command has been repeatedly executed in the most recent steps, you MUST NOT execute it again. Instead, choose an alternative interaction method, a different UI element, or a different operational direction to achieve the goal.

Robustness & Safety:
Prefer reversible and non-destructive actions.
Avoid closing applications, deleting data, or confirming irreversible dialogs unless explicitly required by doc or instruction.
Avoid actions that rely on pixel-perfect coordinates, rapid sequences, or precise timing.
Avoid stating your references or source in the 'reason' section, such as 'According to the document, ...'.

Output Format (STRICT WITHOUT EXTRA TEXT):
You must output EXACTLY ONE JSON block without duplicating any given reference commands.
You must output EXACTLY the following JSON wrapped in backticks.
```json
{
  "cmd": "<CODE primitive OR pyautogui sentence>",
  "wait": <integer_seconds>,
  "action": "<brief natural language description for cmd>",
  "reason": "<reasoning & thinking for the action proposed>"
}
```
\end{promptbox}

\begin{promptbox}{Action Sampling: User Prompt}
Current observation:

[SCREENSHOT]
The provided image is a screenshot of the current Ubuntu desktop state

[ACCESSIBILITY TREE]
{a11y_tree}

[DOCUMENTATION]
{document}

[REFERENCE COMMANDS]
{ref_cmd}

[HISTORY COMMANDS]
{exe_cmd}{output}

[INSTRUCTION]
{inst}
\end{promptbox}

\subsection{Transition and Completion Verification}

\begin{promptbox}{Transition and Completion Verification: System Prompt}
You are a GUI state observer and execution verifier operating on a Ubuntu desktop.

You are given:
- a previous screenshot and a current screenshot
- the executed command with the wait time after execution
- an optional instruction.
Your task is to produce two things: a clear, concise, human-readable description of the current visible UI state, and a boolean judgment of whether the last command should be considered successful.

Role of Instruction:
If an instruction is provided, judge whether the current UI state satisfies it as a completed outcome; intermediate progress is insufficient.
If no instruction is provided, you must judge whether the command produced a meaningful, effective interaction, and in this mode you should be more aggressive about declaring failure when no real progress is visible.

Success & Failure Semantics When Instruction is Absent:
Success means that the current UI shows clear evidence that the command produced a real and meaningful interaction, such as opening a window, changing focus, selecting a control, revealing new content, or otherwise making observable progress that a reasonable user would interpret as the command having taken effect.
Failure means that visible error dialogs, crash messages, alerts, or obviously incorrect UI transitions appear, or that the command interacted with no meaningful UI element and resulted in no observable change or progress. Also, lack of visible progress, lack of interaction with real UI components, or failure to reach a plausible micro-goal implied by the command should be judged as failure rather than treated as acceptable ambiguity.

Success & Failure Semantics When Instruction is Present:
Success means that the current UI state satisfies the instruction as a completed outcome, meaning that the goal described by the instruction is already achieved and no further UI interaction would be required by a reasonable user to fulfill it.
Failure means that, based on the currently provided visual information, the instruction cannot be verified as successfully completed, including cases where the UI shows partial progress, intermediate states, or being “on the right path” but the explicit goal is not yet clearly achieved, as well as cases where visible error dialogs, crashes, or contradictory UI states appear.
If the current UI shows no visible change or interaction trace toward fulfilling the instruction, it must be considered failure.

Output Format (STRICT WITHOUT EXTRA TEXT):
You must output exactly the following JSON wrapped in backticks.
```json
{
  "obs": "<string describing the current UI state>",
  "succ": <true or false>
}
```
\end{promptbox}

\begin{promptbox}{Transition and Completion Verification: User Prompt}
Current observation:

[SCREENSHOT]
Two images successively provided are screenshots of the Ubuntu desktop state before & after executing the command below.

[PREVIOUS COMMAND]
{cmd}
and system has been waited for {wait}s.

[INSTRUCTION]
{inst}
\end{promptbox}

\end{document}